\documentclass[a4paper, 10pt, conference]{ieeeconf}      % Use this line for a4 paper

\IEEEoverridecommandlockouts                              % This command is only needed if 
\pdfoutput=1
\usepackage{graphics} % for pdf, bitmapped graphics files
\usepackage{epsfig} % for postscript graphics files
\usepackage{mathptmx} % assumes new font selection scheme installed
\usepackage{times} % assumes new font selection scheme installed
\usepackage{amsmath} % assumes amsmath package installed
\usepackage{amssymb}  % assumes amsmath package installed
\usepackage{algorithm} %format of the algorithm
\usepackage{algorithmic} %format of the algorithm
\usepackage{multirow} %multirow for format of table
\usepackage{booktabs} 
\usepackage{graphicx}
\usepackage{subcaption}
\usepackage{subfig}

\usepackage[style=numeric, backend=biber, sorting=none]{biblatex} % 使用biblatex包
\title{\LARGE \bf
Digital Modeling of Massage Techniques and Reproduction by Massage Robot
}

\author{Yuan Xu$^{\dag}$, Kui Huang$^{\dag}$, Weichao Guo and Leyi Du% <-this % stops a space
\thanks{*This work is supported by the Shanghai Pudong Municipal Health Commission (Proj. No. PDZY-2023-1101). Yuan Xu is with the SJTU Paris Elite Institute of Technology, Shanghai Jiao Tong University, Shanghai 200240, China. Weichao Guo is with the State Key Laboratory of Mechanical System and Vibration, School of Mechanical Engineering, Shanghai Jiao Tong University, Shanghai 200240, China and also with Meta Robotics Institute, Shanghai Jiao Tong University, Shanghai 200240, China. Kui Huang and Leyi Du are with school of Gongli Hospital Medical Technology, University of Shanghai for Science and Technology. (Corresponding author: Leyi Du, 13801822579@126.com). $^{\dag}$ indicates equal contributions.}% <-this % stops a space
}

\begin{document}

\maketitle
\thispagestyle{empty}
\pagestyle{empty}

%%%%%%%%%%%%%%%%%%%%%%%%%%%%%%%%%%%%%%%%%%%%%%%%%%%%%%%%%%%%%%%%%%%%%%%%%%%%%%%%
\begin{abstract}

This paper explores the digital modeling and robotic reproduction of traditional Chinese medicine (TCM) massage techniques. We adopt an adaptive admittance control algorithm to optimize force and position control, ensuring safety and comfort. The paper analyzes key TCM techniques from kinematic and dynamic perspectives, and designs robotic systems to reproduce these massage techniques. The results demonstrate that the robot successfully mimics the characteristics of TCM massage, providing a foundation for integrating traditional therapy with modern robotics and expanding assistive therapy applications.

\end{abstract}

%%%%%%%%%%%%%%%%%%%%%%%%%%%%%%%%%%%%%%%%%%%%%%%%%%%%%%%%%%%%%%%%%%%%%%%%%%%%%%%%
\section{Introduction}

Traditional Chinese massage (TCM) is based on principles of traditional Chinese medicine, including the concepts of zang-fu organs and meridians. It uses manual techniques to unblock meridians, promote qi and blood circulation, relieve pain, and balance yin and yang. Due to a shortage of TCM practitioners, with an estimated need for 6.25 million more, there is an urgent demand for massage rehabilitation robots that combine intelligent robotics with TCM practices. These robots can provide professional massage therapy and neuro-sensory recovery, improving patients' physical function and quality of life \cite{Yu2021}.

Recent years have seen growing interest in massage robots, with some research focusing on the design of massage hands \cite{Gao2011}, though these are still in experimental stages. Other studies have developed more complete massage rehabilitation systems \cite{Li2012}\cite{Huang2017}, capable of performing various techniques. However, these systems often lack professional medical guidance, do not address massage intensity control, and fail to ensure therapeutic efficacy or safety, limiting their commercial potential.

The human skin and subcutaneous tissues, such as acupoints and muscle groups, serve as key interfaces for human-robot interaction, transmitting displacement and force during massage rehabilitation. Factors like force intensity, duration, and variation directly impact therapeutic efficacy. Current dynamic control methods for massage robots rely on constant force settings, which fail to replicate the complex patterns of traditional Chinese massage techniques. Additionally, due to the lack of professional data and technique analysis, these robots are limited to kinematic reproduction and force control. This study addresses this gap by collecting kinematic and dynamic data from professional therapists and develops a force-compliant control algorithm for a massage robot. The algorithm aims to reproduce the dynamics of professional massage techniques, enhancing the robot's scientific and clinical effectiveness.

\section{Massage Robot Platform}

\begin{figure*}[!htp]
  \centering
  \begin{minipage}{0.41\textwidth}
    \centering
    \includegraphics[height=6cm]{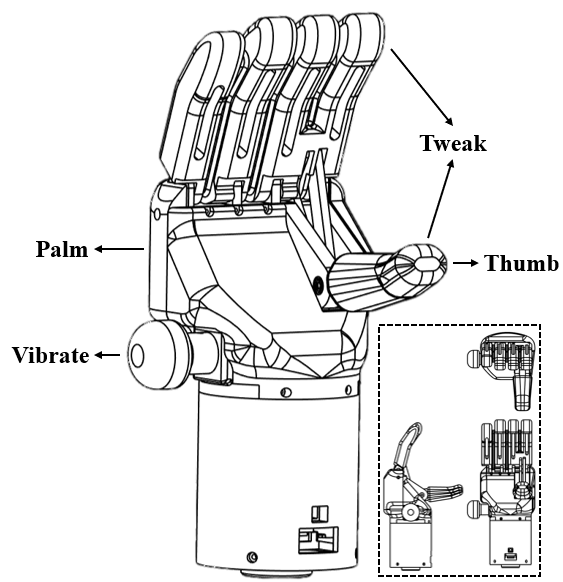}
    \subcaption{Structure of massage hand}
    \label{fig:1-1}
  \end{minipage}
  \hfill
  \begin{minipage}{0.15\textwidth}
    \centering
    \includegraphics[height=6cm]{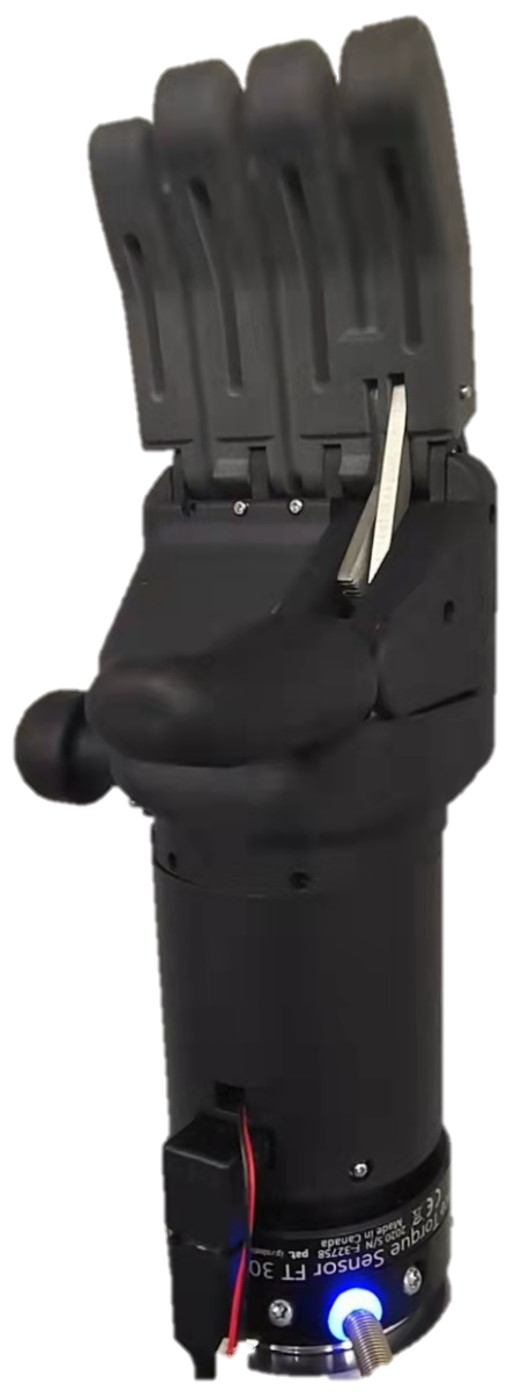}
    \subcaption{Photograph}
    \label{fig:1-2}
  \end{minipage}
  \hfill
  \begin{minipage}{0.41\textwidth}
    \centering
    \includegraphics[height=6cm]{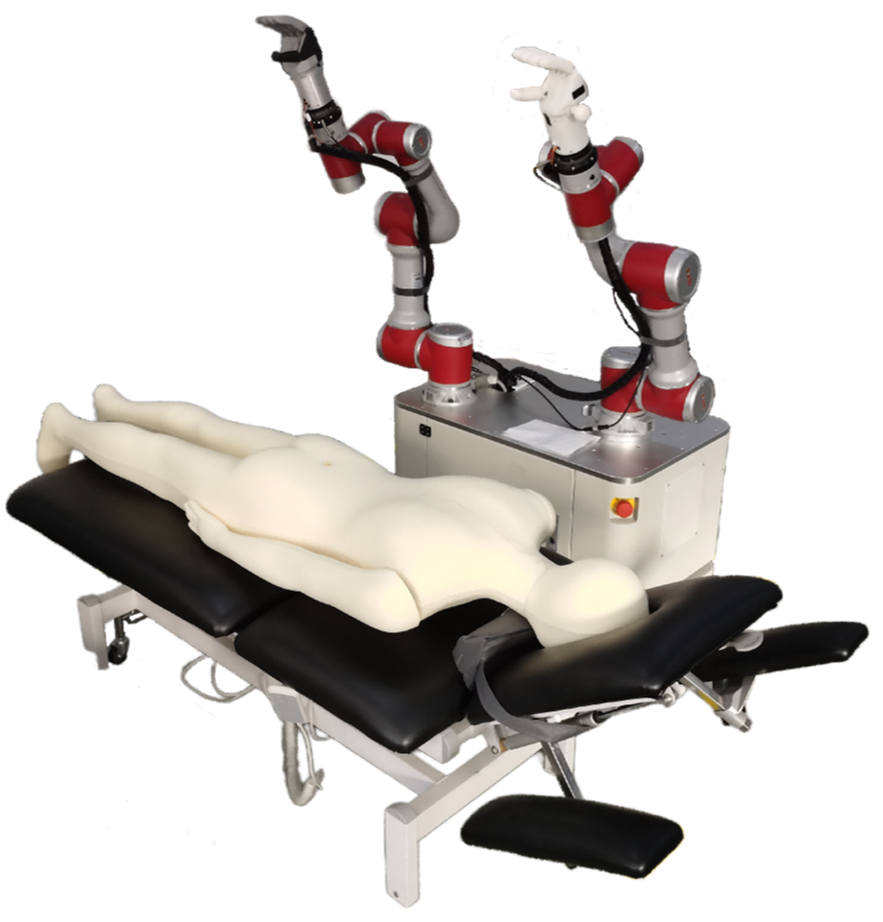}
    \subcaption{Massage robot}
    \label{fig:1-3}
  \end{minipage}
\caption{Massage robot platform}
\label{fig:1}
\end{figure*}

In order to realize a variety of manipulation functions, this paper designs a multifunctional massage hand, as shown in Fig.\ref{fig:1-1} and Fig.\ref{fig:1-2}. Combined with jaka zu7 robotic arm and AGV to build a complete massage robot hardware platform, as shown in Fig.\ref{fig:1-3}.

The design of this mechanical massage hand is inspired by the shape and size of the human palm. The massage hand consists of several core components, including the palm-punch function module, vibration function module, kneading function module, and finger technique function module.

\begin{itemize}
    \item The palm-punch function module is designed based on the natural curvature of the back of the human hand. When the fingers are spread, it performs palm massage, and when the fingers are clenched into a fist, it performs fist massage.
    \item The vibration function module is driven by a vibration motor to simulate high-frequency vibrations, which are used for vibration-based massage.
    \item The kneading function module is driven by a kneading motor, enabling the thumb to pinch with the other four fingers, responsible for kneading massage.
    \item The finger technique function module is designed in the shape of a humanoid thumb and is driven by robotic arm movements to perform finger-based massage techniques.
\end{itemize}

\section{Acquisition and Analysis of Massage Techniques}

\begin{figure*}[!htp]
  \centering
  \begin{minipage}{0.48\textwidth}
    \centering
    \includegraphics[height=5cm]{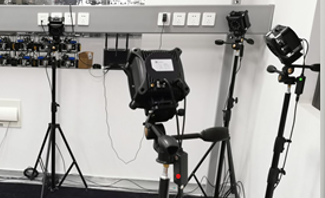}
    \subcaption{Optitrack}
    \label{fig:optitrack}
  \end{minipage}
  \hfill
  \begin{minipage}{0.48\textwidth}
    \centering
    \includegraphics[height=5cm]{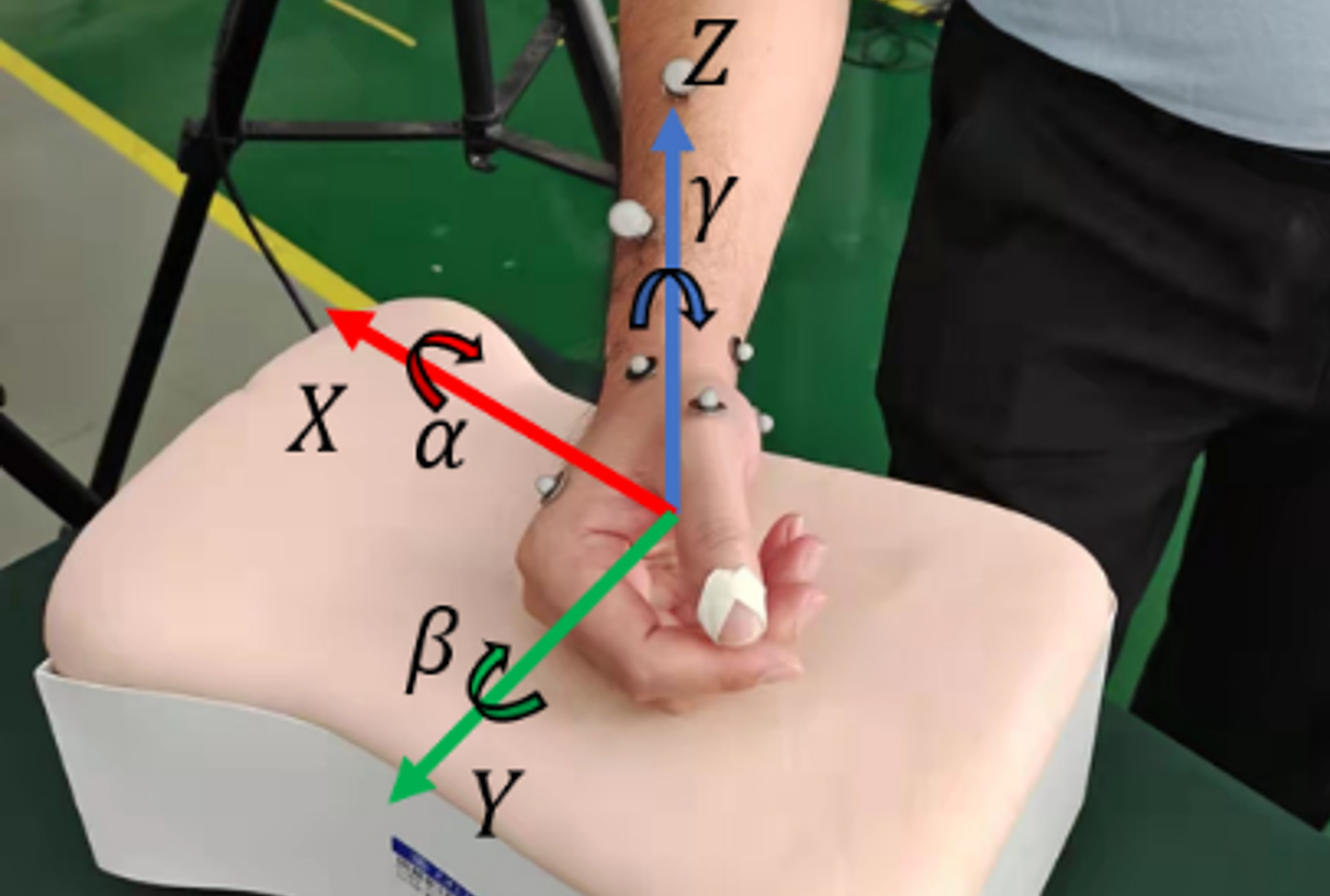}
    \subcaption{Massage axis}
    \label{fig:massage axis}
  \end{minipage}
\caption{Methods of collecting information on expert massage techniques}
\label{fig: methods of collecting}
\end{figure*}

The movements of massage techniques follow certain patterns. This section will provide a kinematic and dynamic analysis of massage techniques, introduce a mechanical massage hand, and discuss the implementation of massage techniques on a dual-arm massage robot.

This study uses the OptiTrack motion capture system and ZTC-II massage strength measuring instrument to collect data on the techniques of massage experts. A coordinate system for the massage techniques, as shown in Fig.\ref{fig: methods of collecting}, is established. By analyzing the expert's technique data, the following kinematic information for the techniques can be obtained:

\begin{enumerate}
    \item \textbf{Beat}: The motion pattern involves both hands forming fists, using the hypothenar eminence to alternately strike the treatment area up and down. The degrees of freedom of the technique are mainly translational motion along $Z$-axis and rotational motion around $\alpha$ . The movement distance is approximately 16.5cm, and the time for one strike with one hand is about $0.45s$. The force direction is along the $Z$-axis translation.

        \begin{figure}[htbp]
        \centerline{\includegraphics[width=0.18\textwidth]{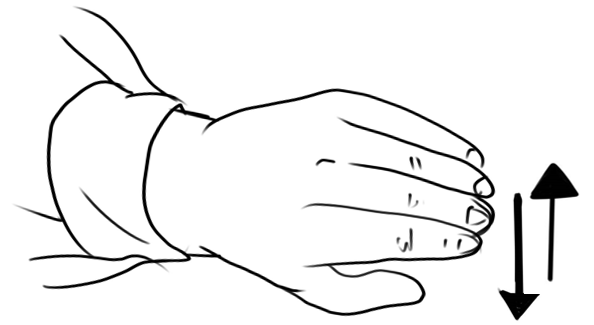}}
        \caption{Description of beat}
        \label{fig:beat_pic}
        \end{figure}

    \item \textbf{Press}: The motion pattern involves applying force with the palm or back of the hand on the target acupoint, gradually pressing down with increasing force, maintaining the pressure for a few seconds, and then releasing. This process is repeated. The degrees of freedom of the technique are primarily translational motion along the $Z$-axis, and the force direction is along the $Z$-axis translation.

        \begin{figure}[htbp]
        \centerline{\includegraphics[width=0.15\textwidth]{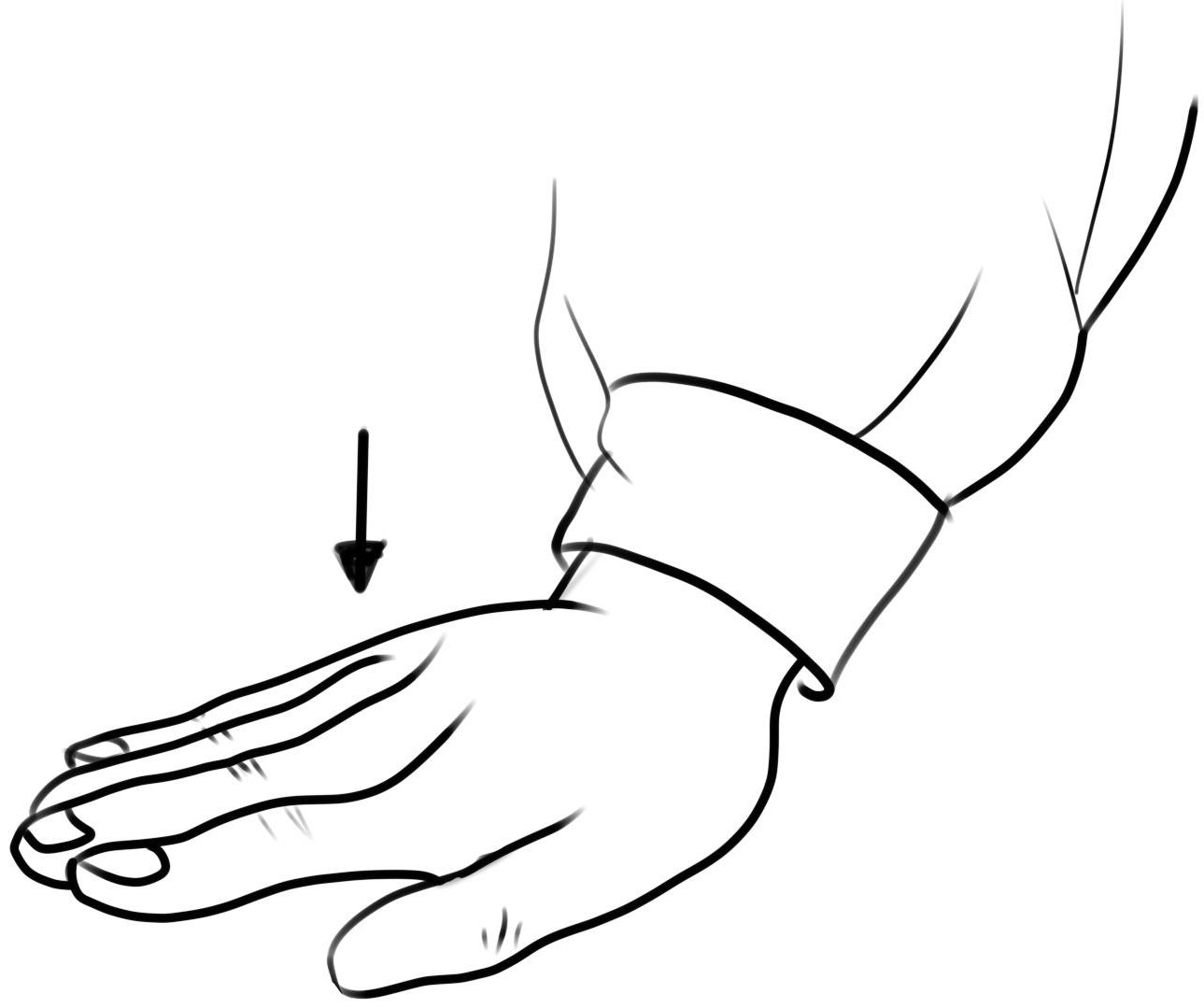}}
        \caption{Description of press}
        \label{fig:press_pic}
        \end{figure}
    
    \item \textbf{Push}:The motion pattern involves fixing the hand at the proximal end, with the palm or back of the main hand placed on the treatment surface. The force is applied through the wrist to cause the palm to sink, performing back-and-forth movements in the forward-backward or left-right directions. The degrees of freedom of the technique are primarily translational motion along the $X$-axis or $Y$-axis. The translation distance for one motion is approximately 20cm, and the force direction is translational motion along the $Z$ and $X$ or $Y$ axes.

        \begin{figure}[htbp]
        \centerline{\includegraphics[width=0.1\textwidth]{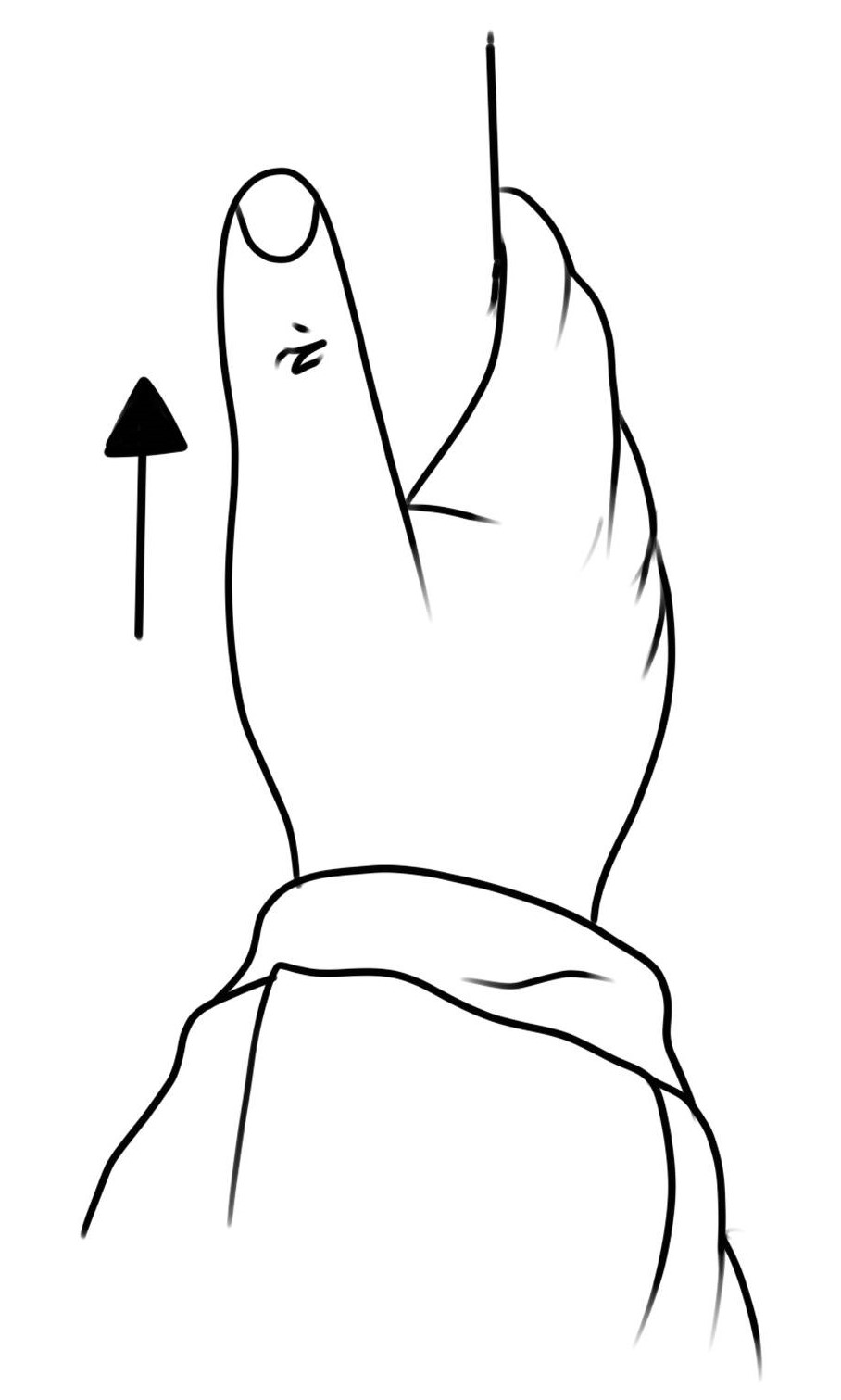}}
        \caption{Description of push}
        \label{fig:push_pic}
        \end{figure}
    
    \item \textbf{Vibrate}: The motion pattern involves applying pressure with the thumb or palm on the target acupoint to generate a rapid vibration wave. The degrees of freedom of the technique are primarily micro-translational motion along the $Z$-axis, and the force direction is micro-translational motion along the $Z$-axis.

        \begin{figure}[htbp]
        \centerline{\includegraphics[width=0.1\textwidth]{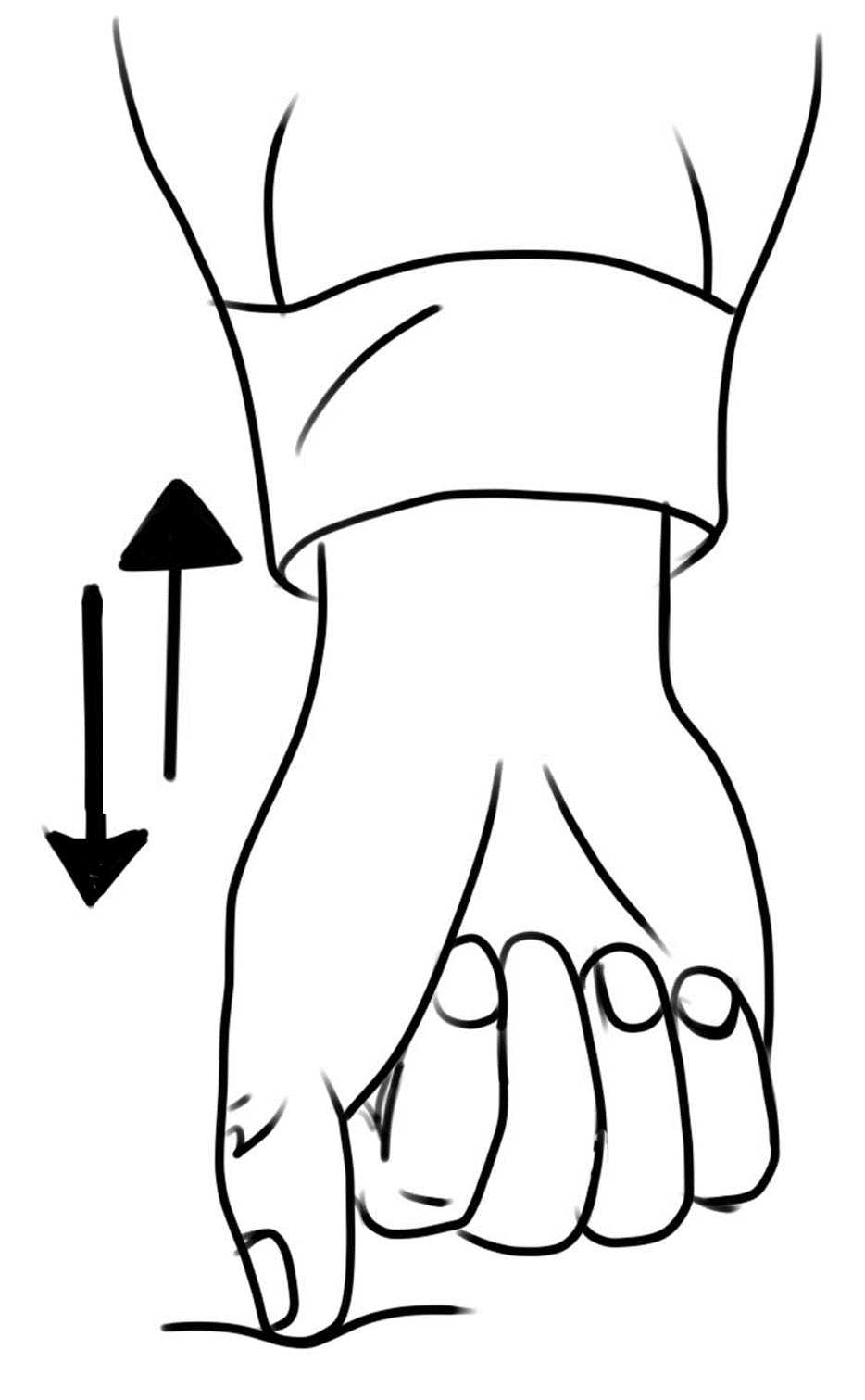}}
        \caption{Description of vibrate}
        \label{fig:vibrate_pic}
        \end{figure}
        
\end{enumerate}

\begin{figure*}[htbp]
\centerline{\includegraphics[width=0.98\textwidth]{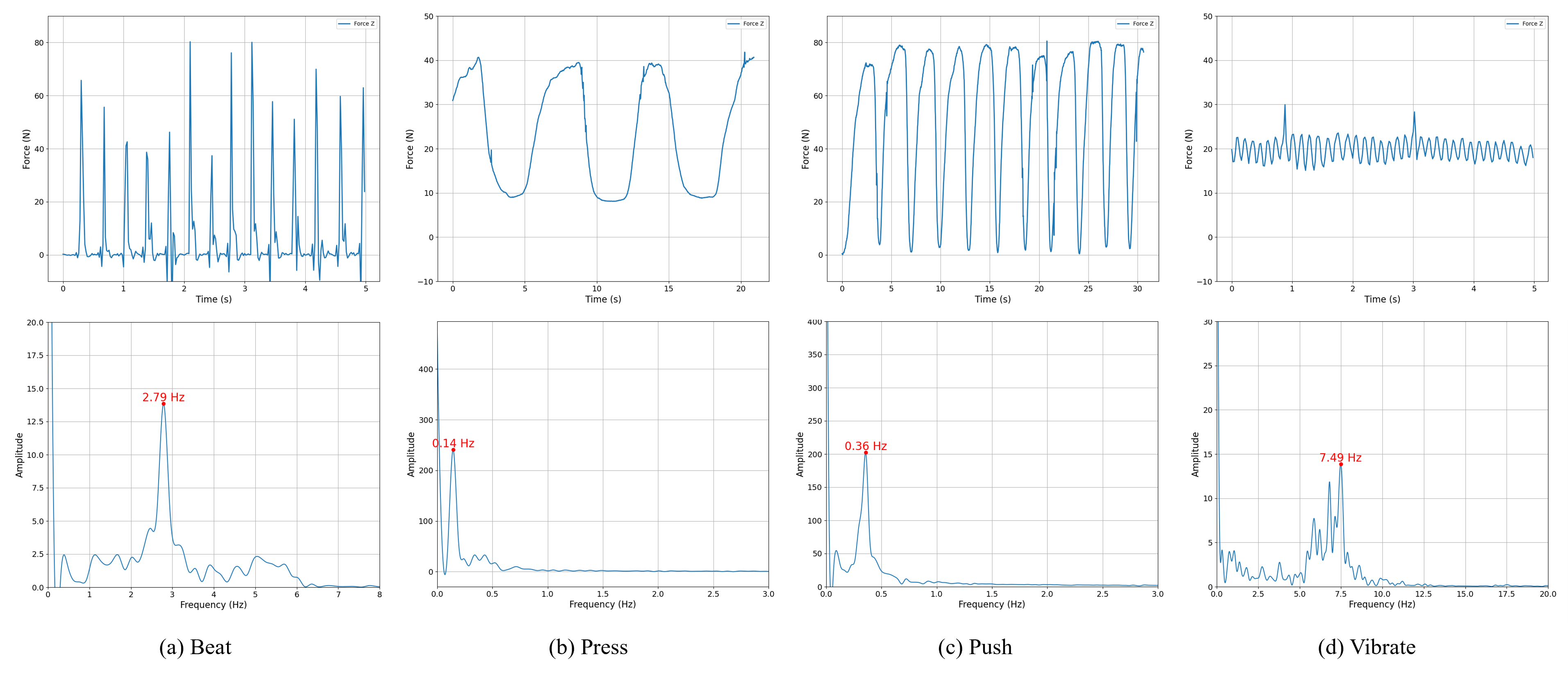}}
\caption{Raw data on the strength and spectrum of the massage technique}
\label{fig:massage_real}
\end{figure*}

After collecting the force data from the expert, the force curves for each massage technique are first plotted using Python, as shown as in Fig.\ref{fig:massage_real}. 

Then, time-domain and frequency-domain analyses are performed to extract the characteristic information of the techniques, as shown as in Tab.\ref{tab:massage_real}.

\begin{table}[!htbp]
\centering
\caption{Time $\&$ frequency characterization of expert massage technique}
\begin{tabular}{cccccccc}
   \toprule
    Parameter & Beat & Press & Push & Vibrate \\
    \midrule   
    Max (N) & 79.67 & 42.59 & 81.65 & 25.74 \\
    Min (N) & 0.00 & 8.20 & 0.00 & 14.39 \\
    Mean $\mu$ (N) & 64.90 & 15.64 & 28.94 & 20.07 \\
    Std $\sigma$ (N) & 12.17 & 6.67 & 9.02 & 2.80 \\
    Skew $\gamma$ & -4.19 & -0.62 & 0.11 & -0.13 \\
    Kurt $\kappa$ & 21.78 & -1.07 & -0.16 & -0.04 \\
    Frequency $f$ (Hz) & 2.79 & 0.14 & 0.36 & 7.49 \\
    \bottomrule
\end{tabular}
\label{tab:massage_real}
\end{table}

\section{Compliant Control and Robotic Reproduction}

In order to address the nonlinear stiffness characteristics of human skin, this study employs a compliance control method based on adaptive admittance. The specific admittance control law is as follows:

\begin{equation}
m \hat{\ddot{e}}+b(\hat{\dot{e}}+\phi(t))=f_d-f_e
\end{equation}

where $\phi(t)$ is the adaptive compensation term.

\begin{equation}
\phi(t)=\phi(t-T)+\sigma \frac{f_e(t-T)-f_d(t-T)}{b}
\end{equation}

This term adjusts the damping coefficient based on the force error from the previous cycle.

Since the industrial robotic arm provides position control mode, the adaptive admittance control law is converted into a discrete position control mode as follows:

\begin{equation}
\left\{\begin{array}{l}
\ddot{x}_c(t)=\ddot{x}_e(t)+\frac{1}{m}\left[\Delta(f)-b(t)\left(\dot{x}_c(t-1)-\dot{x}_e(t)\right)-\phi(t)\right] \\
\dot{x}_c(t)=\dot{x}_c(t-1)+\ddot{x}_c(t) * T \\
x_c(t)=x_c(t-1)+\dot{x}_c(t) * T
\end{array}\right.
\end{equation}

The stability and convergence of this method have been validated \cite{duan2018adaptive}.

The core of the robot's reproduction of massage techniques lies in adjusting factors such as massage force, frequency, motion direction, and target area. This section explores the implementation of robotic massage techniques based on the aforementioned massage methods, incorporating the suggestions of massage experts.

\begin{enumerate}
    \item \textbf{Beat}: Based on the kinematic and dynamic analysis of the massage technique, the motion of the beat technique primarily consists of the following two movements:
        \begin{enumerate}
            \item Movement 1: The massage hand forms a fist and makes contact with the patient. The robot's end-effector rotates around the X-axis by $25^{\circ}$, while translating $3\text{cm}$ along the X-axis and $16\text{cm}$ along the $Z$-axis. The entire process lasts for $1s$.
            \item Movement 2: This is the inverse of Movement 1. However, to reduce the force error caused by position control inaccuracies in the robotic arm, the angle for Movement 2 is set as follows:
                \begin{equation}
                    r_{x,2} = r_{x,1} + cos(\omega t)+\delta
                \end{equation}
            Where $cos(\omega t)+\delta$ is the correction factor used to compensate for the accumulated mechanical errors of the robot over time.
        \end{enumerate}

    \item \textbf{Press}: Based on the kinematic and dynamic analysis of the massage technique, the press technique corresponds to a desired force oscillating in a sine wave pattern, with the peak and trough held for $3$ seconds and $1$ second, respectively.
        \begin{equation}
            f_{z, \text { desired }}(t)= \begin{cases}A \sin (\omega t) & \text { if } 0 \leq t<T/4 \\ A &\text{for 3s}  \text { if } t = T/4 \\ A \sin (\omega t) & \text { if } T/4<t< 3T/4 \\ -A &\text{for 1s}  \text { if } t = 3T/4 \end{cases}
        \end{equation}
    
    \item \textbf{Push}: Based on the kinematic and dynamic analysis of the massage technique, the push technique involves the reciprocal movement of the back of the hand from the lower back to the hip. The mathematical form of the movement trajectory is as follows:
    
        \begin{equation}
            \left\{\begin{array}{l}
            y=y_0+A\sin (\omega t) \\
            z=z_0+G(f)
            \end{array}\right.
        \end{equation}

    Where $A$ represents the amplitude of the motion, and $G(f)$ represents the displacement along the $Z$-axis generated by the adaptive admittance control.

    \item \textbf{Vibrate}: Based on the kinematic and dynamic analysis of the massage technique, the vibrate technique involves a very high frequency that cannot be achieved solely through the robotic arm's motion. Once the contact force of the massage hand's vibrating head with the patient reaches the desired force, the vibration module of the mechanical massage hand is activated to complete the process.
\end{enumerate}

\section{Experiment and Result}

\begin{figure}[htbp]
\centerline{\includegraphics[width=0.4\textwidth]{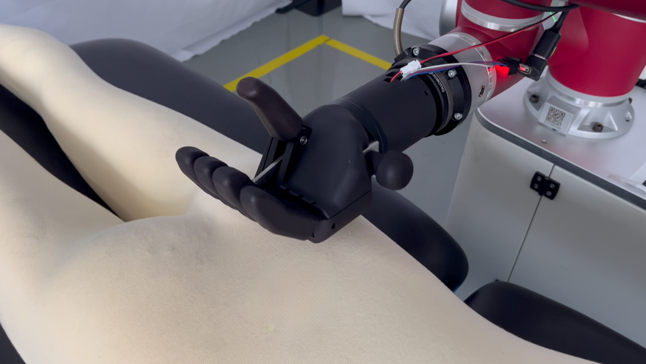}}
\caption{Experimental setup}
\label{fig:exp_setup}
\end{figure}

The robot massage techniques were implemented on a soft mannequin based on the methods described above, and the end-effector contact force was recorded using the Robotiq FT300s force sensor, The experimental setup is shown in the Fig.\ref{fig:exp_setup}.

\begin{figure*}[htbp]
\centerline{\includegraphics[width=0.98\textwidth]{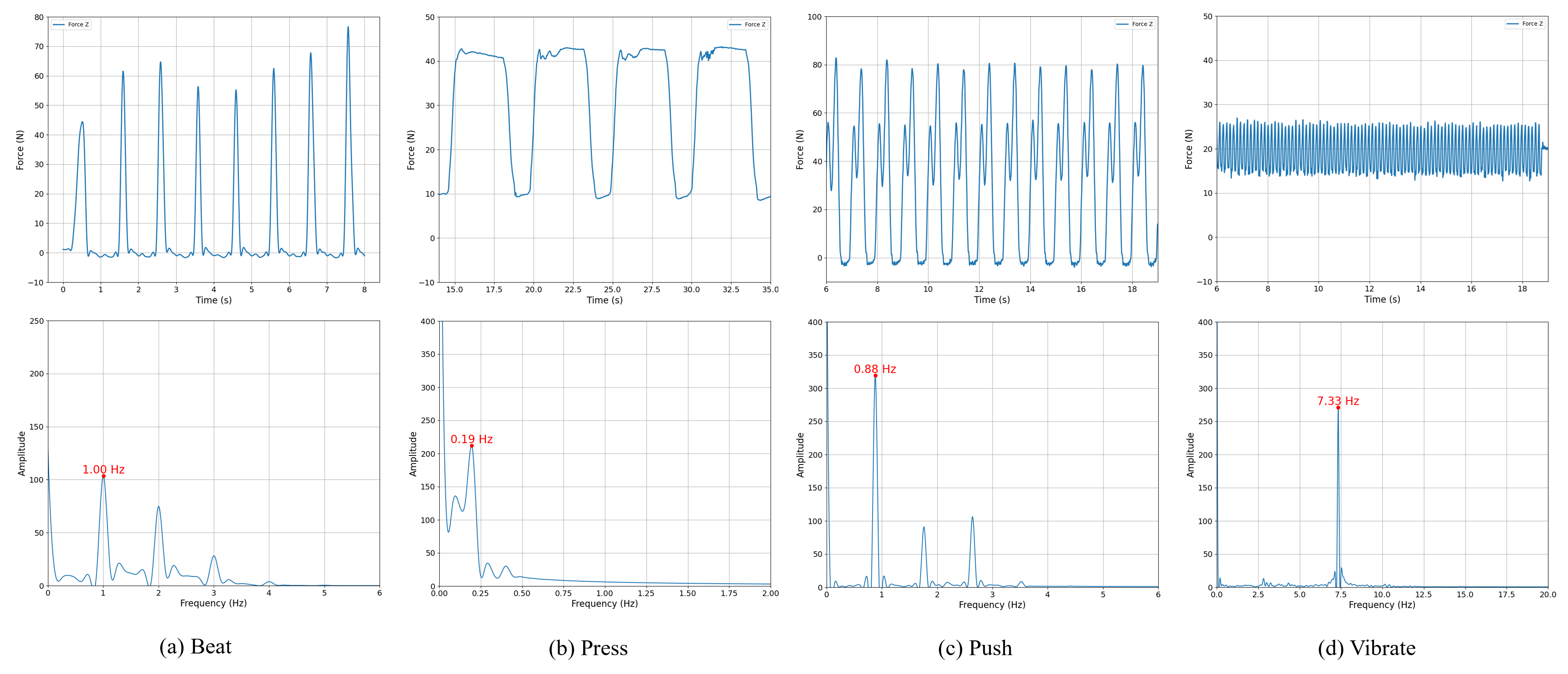}}
\caption{Robotic data on the strength and spectrum of the massage technique}
\label{fig:massage_robot}
\end{figure*}

Fig.\ref{fig:massage_robot} are the force curves and corresponding spectrograms for the robot massage, including the "Beat", "Press", "Push", and "Vibrate" techniques.

\section{Discussion}

This section will compare Fig.\ref{fig:massage_real} and Fig.\ref{fig:massage_robot}, using both qualitative and quantitative methods. First, the robot massage data is analyzed for time-domain and frequency-domain characteristics, and Tab.\ref{tab:massage_robot} is plotted accordingly.

\begin{table}[!htbp]
\centering
\caption{Time $\&$ frequency characterization of robot massage technique}
\begin{tabular}{cccccccc}
   \toprule
    Parameter & Beat & Press & Push & Vibrate \\
    \midrule   
    Max (N) & $\mathbf{76.67}$ & $\mathbf{43.22}$ & $\mathbf{82.83}$ & $\mathbf{26.96}$ \\
    Min (N) & $\mathbf{0.00}$ & $\mathbf{8.50}$ & $\mathbf{0.00}$ & $\mathbf{12.69}$ \\
    Mean $\mu$ (N) & 8.45 & 31.72 & $\mathbf{27.51}$ & $\mathbf{19.16}$ \\
    Std $\sigma$ (N) & $\mathbf{18.02}$ & 13.92 & 27.78 & $\mathbf{4.18}$ \\
    Skew $\gamma$ & 1.99 & $\mathbf{-0.78}$ & $\mathbf{0.34}$ & $\mathbf{0.09}$ \\
    Kurt $\kappa$ & $\underline{2.73}$ & $\mathbf{-1.24}$ & $\mathbf{-1.26}$ & $\mathbf{-1.52}$ \\
    Frequency $f$ (Hz) & $\underline{1.00}$ & $\mathbf{0.19}$ & $\mathbf{0.88}$ & $\mathbf{7.33}$ \\
    \bottomrule
\end{tabular}
\label{tab:massage_robot}
\end{table}

By analyzing the data in the images and tables, and verifying the effectiveness of each massage technique, it can be observed that all techniques replicate the expert's massage force well. However, there are slight differences in terms of force variation trends and frequency. Below is a detailed analysis of each technique:

\begin{enumerate}
    \item \textbf{Beat}:The force curve of the "Beat" technique shows significant fluctuations, reflecting the strong variation and rebound characteristics of the strike. With a skewness of 1.99 (indicating right skew) and a kurtosis of 2.73, the robot has captured the main features of the striking motion, although its kurtosis and frequency (1.00 Hz) are lower than the expert's (21.78 and 2.79 Hz). Overall, the mechanical performance meets the expected results of the "Beat" technique.

    \item \textbf{Press}:The force curve of the "Press" technique shows significant fluctuations, simulating the strong variation in force during pressing and consistently achieving peak compression. With a skewness of -0.78 (indicating left skew) and a kurtosis of -1.24, the force distribution is more stable with fewer outliers. The characteristic frequency is 0.19 Hz, slightly higher than the expert's frequency. Overall, the mechanical performance aligns with the expected outcome of the "Press" technique.

    \item \textbf{Push}:The force curve of the "Push" technique shows significant fluctuations, simulating the repeated pushing motion between the waist and hips. With a skewness of 0.34 (slightly right-skewed) and a kurtosis of -1.26, the force distribution is relatively smooth with few outliers. The characteristic frequency is 0.88 Hz, much higher than the expert's frequency, but the robot still effectively replicates the force variation pattern of the "Push" technique. Overall, the mechanical performance meets the expected outcome of the "Push" technique.

    \item \textbf{Vibrate}:The force curve of the "Vibrate" technique shows small fluctuations with frequent minor changes, reflecting its gentle nature. The skewness is 0.09, indicating a nearly symmetric force distribution, while the kurtosis of -1.52 suggests a smooth distribution with few outliers. The characteristic frequency is 7.33 Hz, similar to the expert's frequency. Overall, the mechanical performance aligns with the expected outcome of the "Vibrate" technique.
\end{enumerate}

In conclusion, the robot can effectively replicate the four massage techniques: Beat, Press, Push, and Vibrate.

\section{Conclusion}

In conclusion, this paper mainly explores the key issues in the digital modeling of massage techniques and their reproduction by massage robots within a robotic massage system, and proposes solutions. We adopted a compliance control algorithm based on adaptive admittance control, aiming to optimize force and position control to address challenges such as patient posture changes and muscle stiffness differences, ensuring the safety and comfort of the massage process. Furthermore, combining massage techniques, this paper conducts an in-depth analysis of several typical massage methods from the kinematic and dynamic perspectives, designs corresponding mechanical massage hands, and uses robotic technology to replicate and automate these traditional techniques. Through the introduction of these technologies, this paper provides theoretical support and practical guidance for the robotic implementation of massage techniques, promoting the integration of traditional Chinese massage with modern robotics, and opening up new application prospects for assistive therapeutic methods.

\addtolength{\textheight}{-12cm}   % This command serves to balance the column lengths

%%%%%%%%%%%%%%%%%%%%%%%%%%%%%%%%%%%%%%%%%%%%%%%%%%%%%%%%%%%%%%%%%%%%%%%%%%%%%%%%

%%%%%%%%%%%%%%%%%%%%%%%%%%%%%%%%%%%%%%%%%%%%%%%%%%%%%%%%%%%%%%%%%%%%%%%%%%%%%%%%

%%%%%%%%%%%%%%%%%%%%%%%%%%%%%%%%%%%%%%%%%%%%%%%%%%%%%%%%%%%%%%%%%%%%%%%%%%%%%%%%

%%%%%%%%%%%%%%%%%%%%%%%%%%%%%%%%%%%%%%%%%%%%%%%%%%%%%%%%%%%%%%%%%%%%%%%%%%%%%%%%

% \bibliographystyle{IEEEtran}
\printbibliography

\end{document}